%% file: main.tex
\documentclass[10pt,twocolumn,letterpaper]{article}

\usepackage[preprint]{iccv}      


\definecolor{iccvblue}{rgb}{0.21,0.49,0.74}
\usepackage[pagebackref,breaklinks,colorlinks,allcolors=iccvblue]{hyperref}
\usepackage{algorithm}
\usepackage{algpseudocode}

\usepackage{tabularx}

\def\paperID{11402} 
\def\confName{ICCV}
\def\confYear{2025}

\title{Seeing and Reasoning with Confidence: Supercharging Multimodal LLMs with an Uncertainty-Aware Agentic Framework}

\author{
Zhuo Zhi\textsuperscript{1}, Chen Feng\textsuperscript{1}, Adam Daneshmend\textsuperscript{2}, Mine Orlu\textsuperscript{1}, Andreas Demosthenous\textsuperscript{1},\\
Lu Yin\textsuperscript{3}, Da Li\textsuperscript{4}, Ziquan Liu\textsuperscript{5}, Miguel Rodrigues\textsuperscript{1}\\[1ex]
\textsuperscript{1}University College London, UK\\
\textsuperscript{2}National Health Service, UK\\
\textsuperscript{3}University of Surrey, UK\\
\textsuperscript{4}Samsung AI Centre, UK\\
\textsuperscript{5}Queen Mary University of London, UK
}
\begin{document}
\maketitle
\footnotetext[0]{Preprint, Under review.}
\input{sec/0_abstract}    
\input{sec/1_intro}
\input{sec/2_related_work}
\input{sec/3_method}

\input{sec/4_experiment}

\input{sec/5_conclusion}

{
    \small
    \bibliographystyle{ieeenat_fullname}
    \bibliography{main}
}
\input{sec/6_appendix}

\end{document}

%% file: sec/0_abstract.tex
\begin{abstract}
Multimodal large language models (MLLMs) show promise in tasks like visual question answering (VQA) but still face challenges in multimodal reasoning. Recent works adapt agentic frameworks or chain-of-thought (CoT) reasoning to improve performance. However, CoT-based multimodal reasoning often demands costly data annotation and fine-tuning, while agentic approaches relying on external tools risk introducing unreliable output from these tools. In this paper, we propose Seeing and Reasoning with Confidence (SRICE), a training-free multimodal reasoning framework that integrates external vision models with uncertainty quantification (UQ) into an MLLM to address these challenges. Specifically, SRICE guides the inference process by allowing MLLM to autonomously select regions of interest through multi-stage interactions with the help of external tools. We propose to use a conformal prediction-based approach to calibrate the output of external tools and select the optimal tool by estimating the uncertainty of an MLLM's output. Our experiment shows that the average improvement of SRICE over the base MLLM is 4.6$\%$ on five datasets and the performance on some datasets even outperforms fine-tuning-based methods, revealing the significance of ensuring reliable tool use in an MLLM agent.
\end{abstract}
\vspace{-0.6cm}

%% file: sec/1_intro.tex
\section{Introduction}
\label{sec:intro}
\begin{figure}
    \centering
    \includegraphics[width=0.9\linewidth]{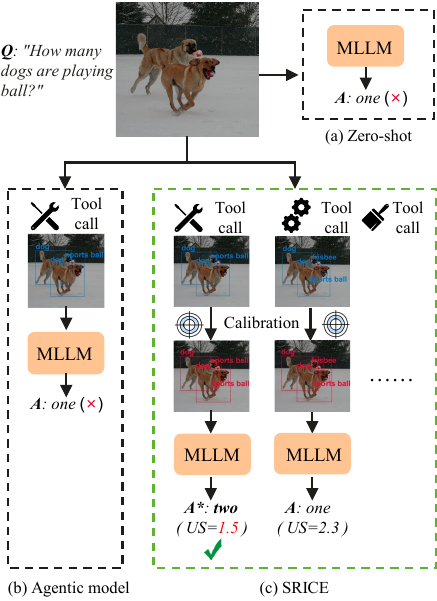}
    \vspace{-0.2cm}
    \caption{Comparison of different multimodal reasoning methods on an example image. \emph{US} refers to the uncertainty score. (a) Zero-shot VQA by the MLLM. (b) The MLLM calls an object detector for additional visual information. (c) Our proposed SRICE framework. The MLLM alone fails in zero-shot mode due to its limited vision recognition capacity. When relying on external models, the output can be unreliable and lead to an incorrect answer. In contrast, SRICE calibrates the outputs from external tools to ensure reliable visual information and estimates the uncertainty of MLLM outputs to select the most reliable tool, yielding the correct answer.}
    \vspace{-0.75cm}
    \label{fig:detection_fail}
\end{figure}
\begin{figure*}
    \centering
    \vspace{-0.6cm}
    \includegraphics[width=1\linewidth]{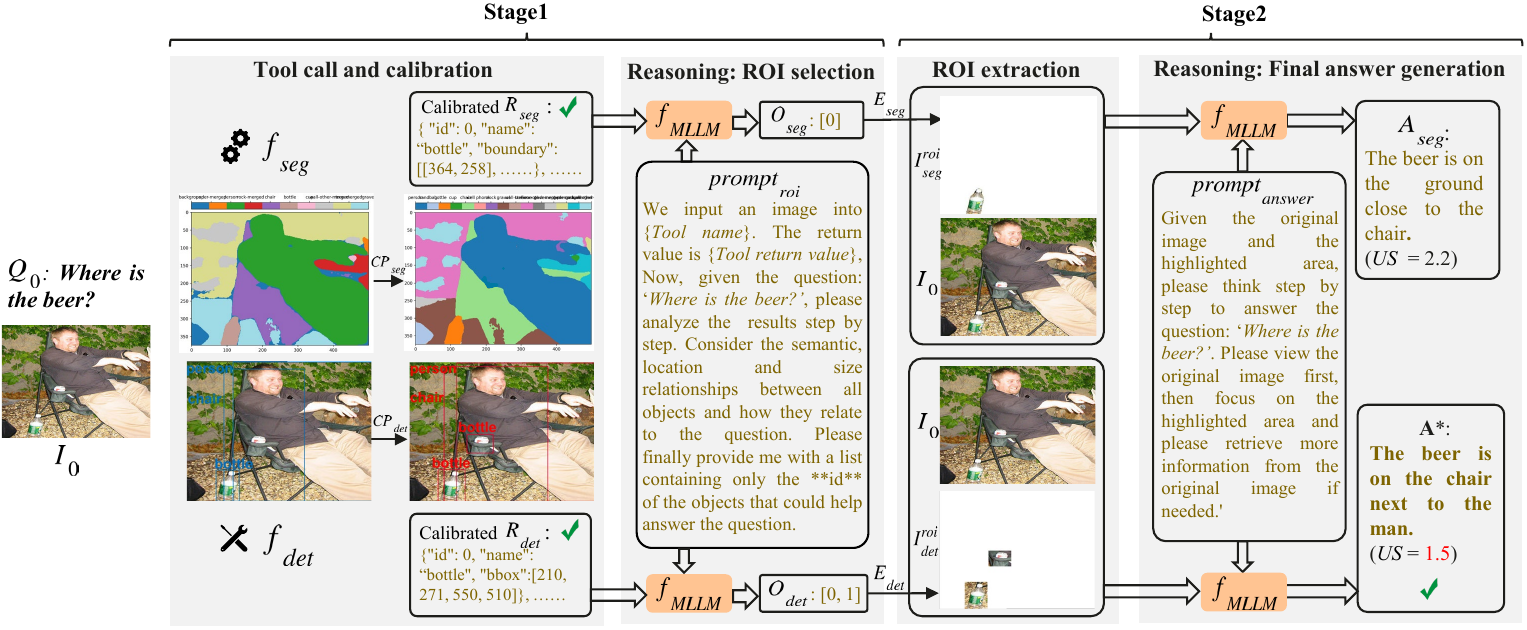}
    \vspace{-0.6cm}
    \caption{The proposed SRICE framework. The MLLM gives the wrong answer 'The beer is on the ground.' for this example. Instead, SRICE  generates the correct answer through a two-stage process. \textbf{Stage 1:} SRICE calls external tools to obtain fine-grained information and applies CP-based calibration to their outputs to improve quality. This calibration mitigates issues such as the segmentation tool misclassifying many pixels as background and the object detection tool missing small objects. Based on the calibrated results, the MLLM selects regions of interest through a CoT process. \textbf{Stage 2:} The key area identified in Stage 1 is extracted and combined with the original image as the MLLM input to perform CoT reasoning. The best answer is chosen from all agentic pathways using our uncertainty estimation based on the prediction set size. } 
    \label{fig:main_fig}
    \vspace{-0.4cm}
\end{figure*}

To mitigate the shortcomings of multimodal large language models (MLLMs) in multimodal reasoning tasks, such as visual question answering (VQA), recent works design agentic frameworks \cite{yao2023react} or multimodal chain-of-thought (CoT) reasoning \cite{wei2022chain} to improve their performance \cite{xu2024llava,shao2024visual}. For instance, MLLMs such as LLaVA \cite{liu2024visual} and Qwen-VL \cite{bai2023qwen} struggle with complex spatial reasoning, Visual-CoT \cite{shao2024visual} and LLaVA-o1\cite{xu2024llava} are proposed to use high-quality reasoning data to fine-tune MLLMs for better multimodal reasoning. On the other hand, image-of-thought (IoT) \cite{zhou2024image} is proposed to use external visual models, such as object detection and segmentation, to provide more accurate visual information for MLLMs without reasoning data collection and fine-tuning. Despite their success, there remain key challenges in existing multimodal CoT and agentic methods. Multimodal CoT's performance needs high-quality training data and fine-tuning, resulting in high-cost in both data collection and computation. Tool-use methods are training-free but when external models are used, the overall performance is heavily dependent on the output quality of external models.

Our paper addresses the limitations of the two existing approaches by proposing a multimodal reasoning framework, \emph{Seeing and Reasoning with Confidence (SRICE)}, which operates with low overhead while leveraging external functions with controlled reliability.  Similar to IoT, SRICE incorporates  external visual models to enhance  the visual perception of an MLLM. In contrast to  Visual-CoT\cite{shao2024visual}, which  explicitly trains the model to find regions of interest, our method enables the MLLM to autonomously identify these regions, thereby facilitating the CoT process. This framework  alleviates the high costs associated with data collection and fine-tuning. The key contribution of SRICE is to address the reliability issue of both external models and the MLLM itself by incorporating uncertainty quantification (UQ) into the agentic multimodal reasoning framework. The output of external models is not always helpful due to their limited generalization. Consequently, the final output of MLLM with the external model can be unreliable. See Fig.~\ref{fig:detection_fail} for an example of the output of an object detection model that misleads the MLLM. Furthermore, when multiple visual models are available, which model to trust remains a question. 

We investigate the possibility of resolving this key challenge using UQ in the proposed multi-stage SRICE framework, as illustrated in Fig. \ref{fig:main_fig}. In the first stage of SRICE, the MLLM calls external visual models to generate fine-grained visual information based on the query and the input image, where the external model's output is calibrated and fed into the MLLM for it to autonomously select regions of interest. For computational efficiency, we use conformal prediction (CP) \cite{vovk2005algorithmic} as the UQ approach for external vision models. In the second stage of SRICE, given the calibrated fine-grained visual information and the original image, the MLLM generates an answer through the CoT reasoning, whose uncertainty is estimated by a prediction set size-based method that leverages the number of plausible tokens in top-$p$ sampling \cite{Holtzman2020The}. Meanwhile, the final output uncertainty helps select the most reliable vision model for this reasoning task.  Our main contributions are as follows.
\begin{itemize}
    \item  We identify and address the two key challenges in existing multimodal reasoning pipelines, massive data annotation and unreliable tool use, by introducing a multi-stage reasoning framework that calls vision tools with uncertainty calibration and selects which tool to rely on based on the MLLM output uncertainty. 
    \item We propose to use conformal prediction for calibrating the external visual tools and MLLM output to ensure that the two stages of MLLM inference are reliable. The strengths of the two UQ methods over existing UQ methods such as confidence scores and entropy are validated in our empirical study.
    \item  We evaluate our proposed method on five datasets, demonstrating that the performance of MLLM is significantly improved with an average  improvement of 4.6$\%$, which is on par with some methods that require additional training and annotations. Our work demonstrates the importance of reliable output of external tools in a multimodal reasoning agent and provides a promising solution to this problem.
\end{itemize}
We review related work in Sec.\ref{sec:related_work} and describe the proposed SRICE framework in Sec.\ref{sec:method}. Sec.\ref{sec:exp} presents the experimental settings and results, and we conclude our work in Sec.\ref{sec:con}.




%% file: sec/2_related_work.tex
\section{Related work}
\label{sec:related_work}
\noindent\textbf{Vision 
language models}
Vision language models integrate textual and visual representations to enable complex reasoning tasks and  has  always been  the  focus of much attention in the multimodal field \cite{zhang2024vision, zhi2024borrowing, zhi2025wasserstein}. Early approaches primarily focused on joint embedding strategies and attention mechanisms, where images and text are projected into a shared feature space for downstream tasks such as visual question answering, image captioning, and cross-modal retrieval \cite{lu2019vilbert, kim2021vilt}.  With the advent of LLM \cite{achiam2023gpt, touvron2023llama}, researchers have explored ways to merge massive textual encoders/decoders with robust vision encoders, aiming to leverage LLMs’ powerful linguistic reasoning while preserving rich visual features. One line of work investigates contrastive pre-training strategies, popularized by CLIP \cite{radford2021learning}. Another family of methods explores fusion architectures—such as Flamingo \cite{alayrac2022flamingo} and BLIP-2 \cite{li2023blip}—where vision encoders are tightly coupled to transformer-based language models to achieve few-shot or even zero-shot performance on tasks ranging from image captioning to visual grounding. More recent efforts push the boundaries by aligning large language models with visual embeddings, named  MLLM, such as LLaVA \cite{liu2024visual}, Qwen-VL\cite{bai2023qwen} and InternVL \cite{chen2024expanding}.  In our work, we  mainly focus on reliably incorporating  external vision models to enhance the perception and reasoning capabilities of MLLM.


\noindent\textbf{Chain-of-thought in visual reasoning.} The success of CoT in LLM is because it significantly improves the model's reasoning ability. With CoT, instead of answering questions directly, LLM simulates the reasoning of human beings and explicitly performs step-by-step reasoning to improve the performance in complex reasoning tasks \cite{wei2022chain, zhang2022automatic}. Many works have begun to try to apply CoT in MLLM to enhance visual reasoning capabilities. Z. Z et al. \cite{zhang2023multimodal} propose a multimodal-CoT that incorporates language  and vision  modalities into a two-stage framework that separates rationale generation and answer inference. In this way, answer inference can leverage better generated rationales that are based on multimodal information. In \cite{shao2024visual}, the authors create a dataset with bounding boxes for question-relevant regions in the image and design a two-stage VQA pipeline. An MLLM fine-tuned on this dataset automatically  locates regions of interest and provides accurate answers by performing CoT with the help of  global and local image details. Similarly, \cite{xu2024llava} introduces the LLavA-O1-100K dataset, which frames VQA at four levels of granularity and applies stage-level beam search. The resulting LLavA-O1 model displays strong reasoning capabilities. However, these methods rely on extensive manual annotations and additional training, which pose practical challenges as MLLMs grow larger. In our approach, we incorporate external vision tools to help the MLLM autonomously select key regions, which facilitate the construction of the CoT. This removes the need for manual annotations and extra training.

\noindent\textbf{Uncertainty Quantification in MLLM} The quantification of uncertainty in LLMs has  gained significant attention as a critical component for improving reliability and trustworthiness in AI systems  \cite{xiong2023can, ye2025benchmarking, lin2023generating}. Some recent work has begun to extend it to MLLM. V. K. et al. \cite{kostumov2024uncertainty} employ conformal prediction to quantify the uncertainty of 20+ MLLMs and conclude that models with the highest accuracy may also have the highest uncertainty. However, they only focus on  multiple-choice  VQA task.  To solve the hallucination of MLLM under rare images, Y. F. et al. \cite{fang2024uncertainty} propose to use uncertainty-guided token dropout to mitigate errors arising from visual token misinterpretation. Unlike, in \cite{groot2024overconfidence}, the  verbalized uncertainty of MLLM is estimated via prompting and  net calibration error is calculated  to measure the direction of miscalibration. Z. K. et al. \cite{khan2024consistency} propose using the principle of neighborhood consistency to identify unreliable responses from an MLLM in QA tasks. Most of these approaches aim at black-box MLLM and do not consider uncertainty in MLLM-based agentic systems. In contrast, our SRICE framework estimates the uncertainty of both external tools and the MLLM’s outputs, leading to more trustworthy results.

%% file: sec/3_method.tex
\section{Method}\label{sec:method}
In this section, we present the proposed methodology in two main sections: the proposed SRICE framework and the uncertainty quantification approach for it.
\subsection{The proposed SRICE framework}

We leverage external vision tools, such as segmentation and object detection models, to extract localized information from an image. This information enables the MLLM to autonomously select regions of interest without  additional training. These regions, combined with the original image, are subsequently fed into the MLLM. The MLLM then performs a CoT reasoning—progressing from a coarse interpretation to fine-grained details—to generate the final answer. To ensure the reliability of both the vision models and the MLLM output, we utilize  UQ for calibrating the tools and for selecting the final answer from multiple pathways. The overall framework is depicted in Fig. \ref{fig:main_fig}. For example, given an image  $I_0$ and a question $Q_0$, the SRICE framework generates the answer $A^*$ by following four main steps:
\begin{itemize}
    \item  \textbf{Tool call and calibration.} We call external tools to perceive information of $I_0$, i.e., the  segmentation tool $f_{seg}$ and  object detection tool $f_{det}$ by following popular agent-based methods \cite{zhou2024image,wang2024visiongpt}. Note that in principle SRICE can involve  using any number of tools but we only describe the two for simplicity.
    To ensure the reliability of these tools, we design the calibration algorithm $CP_{seg}$ and $CP_{det}$  based on conformal prediction. Thus, we get the calibrated output of tools by $R_{seg} = CP_{seg}(f_{seg}(I_0))$ and $R_{det} = CP_{det}(f_{det}(I_0))$. Note that $R_{seg}$ and $R_{det}$ are in text format and objects are described using boundary point coordinates and bounding box, respectively.  
     \item  \textbf{Reasoning: RoI selection.} Instead of selecting the key area manually \cite{shao2024visual}, We leave it to the MLLM $f_{MLLM}$ itself to choose the area that may help to answer the question based on the tool output $R_{seg}$ and $R_{det}$ by using the prompt $prompt_{roi}$ shown in Fig.~\ref{fig:main_fig}. 
     In this process, MLLM performs CoT  reasoning to fully explore the fine-grained information provided by tools and give the interested object id in the answer $O_{seg}$ = $f_{MLLM}(R_{seg}, Q_0, prompt_{roi})$ and $O_{det}$ = $f_{MLLM}(R_{det}, Q_0, prompt_{roi})$.
     \item \textbf{RoI extraction}. Based on the object of interest obtained in the previous step, we extract the corresponding region from the original image by $I^{roi}_{seg} = E_{seg}(I_0, O_{seg})$ and $I^{roi}_{det} = E_{det}(I_0, O_{det})$, where $E_{seg}$ refers to the area extraction by pixel for segmentation result and $E_{det}$ is the area extraction by bounding box for  detection result.
     \item \textbf{Reasoning: Final answer generation}.  We follow \cite{shao2024visual} to input  the original image with the key area and guide the MLLM to perform CoT reasoning from  coarse-grained to fine-grained by using the prompt $prompt_{answer}$ as shown in Fig.~\ref{fig:main_fig}. 
     So far we get the final answer for both pathways by $A_{seg}$ = $f_{MLLM}(I_0, Q_0, I^{roi}_{seg}, prompt_{answer})$ and $A_{det}$ = $f_{MLLM}(I_0, Q_0, I^{roi}_{det}, prompt_{answer})$.  In the end, we choose the more trustworthy  answer as the final answer $A^* = \min_{k} US(A_{k})$,  $US$ is the proposed  uncertainty score based on the quasi-conformal prediction that will be introduced in next section. 
\end{itemize}
Through the previous steps, the  SRICE framework provides a more trustworthy answer. We introduce the uncertainty-guided approach for SRICE in the next section.
\vspace{-0.2cm}
\subsection{Uncertainty quantification approach for SRICE}
\vspace{-0.2cm}
External tools can introduce uncertainty, which is a major challenge for tool-use approaches and a primary focus of our SRICE framework. To extract trustworthy information from these tools and to select the most reliable pathway, we perform uncertainty quantification on both the tool return values and the MLLM outputs, using CP. CP is a distribution-free uncertainty quantification framework that provides statistically valid prediction sets with guaranteed coverage probability \cite{vovk2005algorithmic}. We describe our uncertainty quantification methods for external tools and  the MLLM outputs, respectively.

\subsubsection{Calibration of tools by conformal prediction}

Given a pre-trained model, inductive CP constructs \emph{prediction sets} that contain the true label with probability at least $1 - \alpha$. In classification terms, one chooses a \emph{nonconformity score} $s(\cdot)$ to measure how ``unusual'' each predicted label is (relative to the ground truth), then estimates a threshold $\hat{q}_\alpha$ from a held-out \emph{calibration set}. Formally, if $\{(X_i, Y_i)\}_{i=1}^n$ is a calibration set of size $n$, CP produces sets
\begin{align}
        \mathcal{C}(X_{\mathrm{test}}) 
    \;&=\; 
    \bigl\{\,\hat{y}\;\mid\; s(X_{\mathrm{test}}, \hat{y}) \;\le\; \hat{q}_\alpha\bigr\}\\    s.t.\quad \mathbb{P}\!\bigl(Y_{\mathrm{test}} &\in \mathcal{C}(X_{\mathrm{test}})\bigr) \;\ge\; 1 - \alpha,
\end{align}
where $\hat{q}_\alpha$ is the $(1-\alpha)$-quantile of all calibration scores $\{s(X_i, Y_i)\}_{i=1}^n$ and is calculated by 
\begin{equation}
\small
   \hat{q}_\alpha 
   \;=\;
   \mathrm{Quantile}\!\Bigl(\bigl\{\,s(X_i, Y_i)\bigr\}_{i=1}^{n} \,\cup\, \{+\infty\},
        \;\;\frac{\lceil (n+1)\,(1-\alpha)\rceil}{n}\Bigr).
\end{equation}
The extra $\{+\infty\}$ and the ceiling term $\lceil (n+1)(1-\alpha)\rceil / n$ ensure finite-sample coverage guarantees \cite{angelopoulos2021gentle}. For a classification model whose output is a probability vector over classes, a common choice of nonconformity score is
\begin{equation}
    s(X_i, Y_i) \;=\; 1 \;-\; p\bigl(Y_i \mid X_i\bigr),
\end{equation}
i.e., ``one minus the predicted probability assigned to the true class.'' Below we show how CP can be specialized to (a) a segmentation model (pixel-wise) and (b) an object-detection model (bounding-box--wise).

\noindent{\textbf{Calibration of a segmentation tool}} Segmentation models can be viewed as a grid of pixel-level classifiers, each outputting a probability distribution over $K$ classes at every pixel location. A frequent practical issue is that many pixels belonging to foreground objects get labeled as \emph{background} (class $0$), causing under-segmentation \cite{xie2021segformer}. We calibrate the result using a pixel-wise CP approach that allows the re-labeling of certain background pixels. We define the following notions, 
\begin{itemize}
    \item $\mathcal{X}$ be the space of images of dimension $H\times W$;
    \item $\mathcal{Y} = \{0,1,\dots,K\}$ be the set of semantic labels (including background = $0$);
    \item $\{(X_i, Y_i)\}_{i=1}^n$ be the calibration set, where $Y_i(u,v)\in\mathcal{Y}$ is the ground truth at pixel $(u,v)$;
    \item $N = n \times H \times W$ be the total number of pixels in the calibration set.
\end{itemize}

A segmentation model provides a probability vector $\mathbf{p}_{i,u,v} \in [0,1]^{K+1}$ over classes at each pixel $(u,v)$ in image $X_i$. We define the pixel-wise nonconformity score:
\begin{equation}
    s(i,u,v) \;=\; 1 \;-\; p_{\,Y_i(u,v)}(i,u,v),
\end{equation}
i.e., one minus the predicted probability that $(u,v)$ is its true class. Thus we can collect all nonconformity scores from the calibration set:
\begin{equation}
\small
    \mathcal{S}
    \;=\;
    \small\{\,s(i,u,v) \;\Bigm|\; 
        i=1,\dots,n;\;u=1,\dots,H;\;v=1,\dots,W
    \small\}.
\end{equation}
Its $(1-\alpha)$-quantile $\hat{q}_\alpha$ (with finite-sample correction) is
\begin{equation}
\small
    \hat{q}_\alpha
    \;=\;
    \mathrm{Quantile}\!\Bigl(\,\mathcal{S}\,\cup\,\{+\infty\},\;\;
           \frac{\lceil (N+1)(1-\alpha)\rceil}{N}\Bigr).
\end{equation}
For a \emph{test image} $X_{\mathrm{test}}$, at each pixel $(u,v)$ we define the conformal prediction set
\begin{equation}
    \mathcal{C}(u,v) 
    \;=\;
    \bigl\{\,k \in \mathcal{Y}\;\mid\;1 - p_{\,k}(u,v)\;\le\;\hat{q}_\alpha\bigr\}.
\end{equation}
Thus $\mathcal{C}(u,v)$ is all classes whose pixel-wise nonconformity scores are no larger than $\hat{q}_\alpha$. We map $\mathcal{C}(u,v)$ to a single \emph{calibrated label} $\hat{y}_{\mathrm{cal}}(u,v)$ via:
\begin{equation}
\small
\hat{y}_{\text{cal}}(u,v) =
\begin{cases}
\displaystyle
\arg\max_{j \in \mathcal{C}(u,v)}\, p_j(u,v),
& \begin{aligned}[t]
   &\text{if } \mathcal{C}(u,v)\neq\emptyset, \\
   &and \ 0 \notin \mathcal{C}(u,v),
\end{aligned}
\\[6pt]
\displaystyle
\arg\max_{j \in \mathcal{C}(u,v)\setminus\{0\}}\, p_j(u,v),
& \begin{aligned}[t]
   &\text{if } \mathcal{C}(u,v)\neq\emptyset, \\
   &and \ 0 \in \mathcal{C}(u,v),
\end{aligned}
\\[6pt]
0,
& \text{if } \mathcal{C}(u,v)=\emptyset,
\end{cases}
\end{equation}
If $0$ (background) appears in $\mathcal{C}(u,v)$, the pixel can be re-labeled to a foreground class within that set, mitigating over-conservative segmentation.

\noindent{\textbf{Calibration of an object detection tool}} We apply bounding-box--wise conformal prediction \cite{andeol2023confident} to calibrate object detections. This addresses extreme cases where tiny objects may require expanding the predicted box to guarantee covering the true object with high probability. We define the following notions, let
\begin{itemize}
    \item $\mathcal{X}$ be the image space;
    \item $\mathcal{Y}=\{b^k\}_{k=1}^{K}$ be the set of $K$ ground-truth bounding boxes in an image, where each box is given by   $b^k = (x_{\min}^k, y_{\min}^k, x_{\max}^k, y_{\max}^k)$;
    \item $\{(X_i, \{b^k_i\})\}_{i=1}^n$ be the calibration set, with $\{b^k_i\}$ the ground-truth boxes of image $X_i$.
\end{itemize}

A trained detector produces a set of predicted boxes $\{\hat{b}^j_i = (\hat{x}_{\min}^j, \hat{y}_{\min}^j, \hat{x}_{\max}^j, \hat{y}_{\max}^j)\}$ for image $X_i$. We match predicted boxes to ground-truth boxes via an IoU threshold $\tau$. Let $\mathcal{M}_i$ be the set of matched pairs $(\hat{b}^j_i,b^k_i)$ with $\mathrm{IoU}(\hat{b}^j_i,b^k_i) \ge \tau$. For each matched pair, define the \emph{additive} nonconformity score:
\begin{equation}
\small
\begin{aligned}
    s^k_j
    \;=\;
    \bigl(\,\hat{x}_{\min}^j - x_{\min}^k,\;\hat{y}_{\min}^j - y_{\min}^k,\; \\
           x_{\max}^k - \hat{x}_{\max}^j,\;y_{\max}^k - \hat{y}_{\max}^j\bigr),
\end{aligned}
\end{equation}
which measures the coordinate-wise prediction errors. The full set of nonconformity scores is collected across all matched pairs:
\begin{equation}
\mathcal{S} = \bigcup_{i=1}^n \bigcup_{(\hat{{b}}^j_i, {b}^k_i) \in \mathcal{M}_i} \left\{ s_j^k[1],\ s_j^k[2],\ s_j^k[3],\ s_j^k[4] \right\},
\end{equation}
where $s^k_j[m]$ is the $m$th coordinate error. To ensure coverage on all four coordinates, we apply a Bonferroni correction, splitting risk $\alpha$ into $\alpha/4$ for each coordinate. Let $\mathcal{S}_m$ be the $m$th-coordinate errors across all matched pairs from all calibration images. Then for each $m\in\{1,2,3,4\}$,
\begin{equation}
\small
   \hat{q}_{\alpha,m}
   \;=\;
   \mathrm{Quantile}\Bigl(\mathcal{S}_m \cup \{+\infty\},\;\;
       \frac{\lceil (|\mathcal{S}_m|+1)\,(1-\alpha/4)\rceil}{|\mathcal{S}_m|}\Bigr).
\end{equation}
On a test image $X_{\mathrm{test}}$, we conformalize each predicted box $\hat{b}^j$ by expanding it to
\begin{equation}
\begin{aligned}
   \mathcal{C}(\hat{b}^j)
   \;=\;
   \Bigl[
      \hat{x}_{\min}^j \;-\;\hat{q}_{\alpha,1},\;
      \hat{y}_{\min}^j \;-\;\hat{q}_{\alpha,2},\; \\
      \hat{x}_{\max}^j \;+\;\hat{q}_{\alpha,3},\;
      \hat{y}_{\max}^j \;+\;\hat{q}_{\alpha,4}
   \Bigr].
\end{aligned}
\end{equation}
This guarantees that if $\hat{b}^j$ is matched to a true box $b^k$, then $b^k$ remains inside $\mathcal{C}(\hat{b}^j)$ with probability at least $1-\alpha$.

\subsubsection{Uncertainty estimation for MLLM output based on prediction set}
By using the SRICE pipeline, answers $A_{seg}$ and $A_{det}$ are generated through two pathways. We propose an uncertainty quantification method based on a prediction set in the top-$p$ sampling, corresponding to conformal prediction with a fixed non-conformity score threshold. Thus, we call it a quasi-conformal prediction UQ for convenience but note that the top-$p$ sampling does not generate a prediction set with any coverage guarantee. This method measures the number of candidate tokens required to "cover" a predefined probability mass $p$ at each generation step in top-$p$ sampling.  For an output sequence $A = \{ w_1, w_2, \dots, w_T \}$, where $w_i$ denotes the $i$th token, we compute the uncertainty score by two steps:
\begin{itemize}
    \item Token probability sorting. For each token $w_i$, let $\mathcal{P}_i = \left[ p_i^{(1)}, p_i^{(2)}, \dots, p_i^{(V)} \right]$ represent its probability distribution over the vocabulary, sorted in descending order $\left( p_i^{(1)} \geq p_i^{(2)} \geq \dots \geq p_i^{(V)} \right)$.
    \item Cumulative thresholding. We compute the minimal number of tokens $k_i$ needed to exceed a threshold $p\in (0,1 ]$:
    \begin{equation}
    \small
        k_i = \min \left\{ k \;\middle|\; \sum_{j=1}^{k} p_i^{(j)} \geq p \right\}.
    \end{equation}
    Here, $k_i$ reflects the breadth of plausible alternatives for  $w_i$. A large $k_i$ indicates high uncertainty while a small one implies confidence (few tokens dominate the probability mass).
    \item Sequence-level aggregation. The final uncertainty score (US) for $A$ is achieved by calculating the mean $k_i$ across all tokens:
    \begin{equation}
    \small
        US(A) = \frac{1}{T} \sum_{i=1}^{T} k_i.  
    \end{equation}
\end{itemize}

 By applying the proposed quasi-conformal prediction-based uncertainty estimation, we get the uncertainty score of two pathways $US(A_{seg})$  and $US(A_{det})$. The answer with a smaller uncertainty score will be considered as more trustworthy and thus output by the SRICE as the final answer $A^* = \min_{k} US(A_{k})$

%% file: sec/4_experiment.tex
\section{Experiment}\label{sec:exp}
\begin{table*}[h]
\vspace{-0.5cm}
\caption{\textcolor{black}{Comparison with baselines on all datasets. }}
\vspace{-0.2cm}
\label{tab:main_results}
\centering
\begin{tabularx}{1.86\columnwidth}{lcccccc}
\toprule
            Method &  VQA2 &  VizWiz  &   GQA &  Flickr30K &  MMBench &  Average \\
\midrule
     LLaVA-1.5-13B & 80.0 &   53.6 & 63.3 &     62.3 &    67.7 &             65.4 \\
       LLaVA-ov-7B (\textit{base model}) & 79.0 &   53.6 & 64.8 &     61.1 &    72.3 &             66.2 \\
        SPHINX-13B & 78.1 &   52.5 & 62.6 &     60.7 &    66.9 &             64.2 \\
  vicuna-7B-VisCoT & 77.4 &   54.8 & 61.6 &     67.1 &    67.3 &             65.6 \\
   \textbf{SRICE (\textit{ours})} & 82.5 &   \textbf{60.1} & 67.3 &    67.2 &    \textbf{77.1} &             70.8 \\
   \hline
LLaVA-ov-7B-VisCoT (\textit{upper bound}) & \textbf{83.3} &   57.2 & \textbf{69.6} &     \textbf{69.9} &    76.6 &             \textbf{71.3} \\
\bottomrule
\end{tabularx}
\vspace{-0.2cm}
\end{table*}

We first describe the experimental settings, then the results of our method and the baselines across five multimodal reasoning datasets, demonstrating the effectiveness of our approach.
\subsection{Experimental setting}
\noindent{\textbf{Baselines.}} We select the following strong baseline for comparison. 
\begin{itemize}
    \item Base model: LLaVA-OneVision-Qwen2-7b (LLaVa-OV)
    \cite{li2024llava}. We selected this base model to build the SRICE pipeline due to its excellent performance and support for multiple image inputs, which is essential for our implementation. We refer to this baseline as LLaVA-ov-7B.
    \item LLaVA-1.5-13B \cite{liu2024improved}. LLaVA-1.5-13B is a popular  baseline for VQA tasks with an increased parameter count.
    \item Visual CoT (VisCoT) \cite{shao2024visual}. VisCoT is a state-of-the-art (SOTA) approach that fine-tunes an MLLM on a manually curated dataset, enabling it to automatically identify key areas in an image and facilitate CoT reasoning. We fine-tune the LLaVa-OV model on the VisCoT dataset following the procedure in \cite{shao2024visual} and denote the resulting model as LLaVA-ov-7B-VisCoT. Additionally, we compare our method with the vicuna-7B-VisCoT model released in \cite{shao2024visual}.
    \item SPHINX \cite{lin2023sphinx}. SPHINX enhances the MLLM's ability to identify regions of interest and guide CoT reasoning by incorporating multiple visual encoders during training. The base model used in SPHINX is Llama2-13B. 
\end{itemize}
\noindent{\textbf{Datasets.}} We select a diverse set of datasets to comprehensively evaluate our proposed method:
\begin{itemize}
    \item VQA2 \cite{balanced_vqa_v2}:  a large-scale benchmark for open-ended VQA tasks, featuring 107,391 question-answer pairs in the test set that cover diverse topics, including object recognition, counting, and commonsense reasoning.  
    \item VizWiz \cite{gurari2018vizwiz}: a real-world dataset collected from blind or low-vision users, capturing everyday scenarios. The open-ended questions frequently involve tasks such as text recognition, object identification, and scene understanding, with 8,000 question-answer pairs in the test split. 
    \item Flickr30K \cite{plummer2015flickr30k}: supports VQA with open-ended question-answer pairs that emphasize fine-grained object attributes, actions, and contextual scene understanding. The test set comprises 1,546 question-answer pairs.
    \item GQA \cite{hudson2019gqa}: provides 12,578 open-ended questions in the test set, with a focus on the relationships among objects, attributes, and spatial arrangements.
    \item MMBench \cite{liu2024mmbench}: includes 1,784 questions in the test set that span classification, captioning, and VQA tasks. It features both multiple-choice and open-ended questions across diverse visual scenarios. \end{itemize}
\noindent{\textbf{Metric.}} All dataset evaluation scripts use accuracy  as a metric, more details are given in Appendix B.

\noindent{\textbf{Tool selection.}} We choose popular lightweight tools: SEEM \cite{zou2023segment} for segmentation and Yolov11 \cite{yolo11_ultralytics} for object detection. We also try other tools in the ablation study. 

\noindent{\textbf{SRICE setting}} We set $\alpha = 0.1$ for both $CP_{seg}$ and $CP_{det}$,  following prior  work \cite{kumar2023conformal, van2025self}. Empirically, we  set the parameter $p$ to 0.9 for the  prediction set-based uncertainty estimation.  To compute the nonconformity scores for both tools, we use the COCO-2017 validation dataset \cite{lin2014microsoft}, which contains 5,000 images with annotations for both segmentation and object detection. The time consumption for calculating the nonconformity score on the calibration set is 35 minutes and 8 minutes for the segmentation and object detection tools, respectively. For each test image, the average time taken for calibration is about 34ms and 0.2ms, respectively. The hardware and software platforms involved in the experiment are given in Appendix B.

\subsection{Main result}
We present the experimental results of our proposed SRICE method alongside baselines on all datasets in Table \ref{tab:main_results}. As shown in the table, SRICE substantially improves the base model’s performance by 3.5$\%$, 6.5$\%$, 2.5$\%$, 6.1$\%$, and 4.8$\%$ across the five datasets, demonstrating its effectiveness. Notably, VizWiz exhibits the largest performance gain, likely because this dataset contains many blurred images, where SRICE’s tool calibration plays a particularly important role. Furthermore, SRICE outperforms all competing methods except for the upper bound, including some methods with more parameters  (e.g., LLaVA-1.5-13B and SPHINX-13B). Remarkably, SRICE even surpasses the upper bound on VizWiz and MMBench, underlining the significance of reliable tool return values.
\begin{table}[t]
\centering
\vspace{-0.3cm}
\caption{\textcolor{black}{Ablation of UQ approach in SRICE.
}}
\label{tab:ablation_answer_selection}
\vspace{-0.3cm}
\centering
\setlength{\tabcolsep}{1pt} 
\resizebox{1\linewidth}{!}{
\begin{tabular}{lcccccc}
\toprule
            Method &  VQA2 &  VizWiz  &   GQA &  Flickr30K &  MMBench &  Average \\
\midrule
       base model & 79.0 &   53.6 & 64.8 &     61.1 &    72.3 &             66.2 \\
SRICE-seg-base &80.3 &   56.2 & 65.2 &    62.3 &    74.0 &      67.6 \\
SRICE-det-base & 80.7 &  57.3 & 65.9 &    63.4 &    74.7 &       68.4 \\
SRICE-seg-CP & 82.0 &   59.0 &66.6 &    66.1  &   75.5 &    69.8\\
SRICE-det-CP & 81.7 &   58.6 & 66.2 &    65.2 &  75.8 &    69.5 \\
\textbf{SRICE (\textit{ours})} & \textbf{82.5} &   \textbf{60.1} & \textbf{67.3} &    \textbf{67.2} &    \textbf{77.1} &             \textbf{70.8} \\
\bottomrule
\end{tabular}
}
\vspace{-0.4cm}
\end{table}

\begin{figure}
    \centering
    \vspace{-0.3cm}
    \includegraphics[width=1\linewidth]{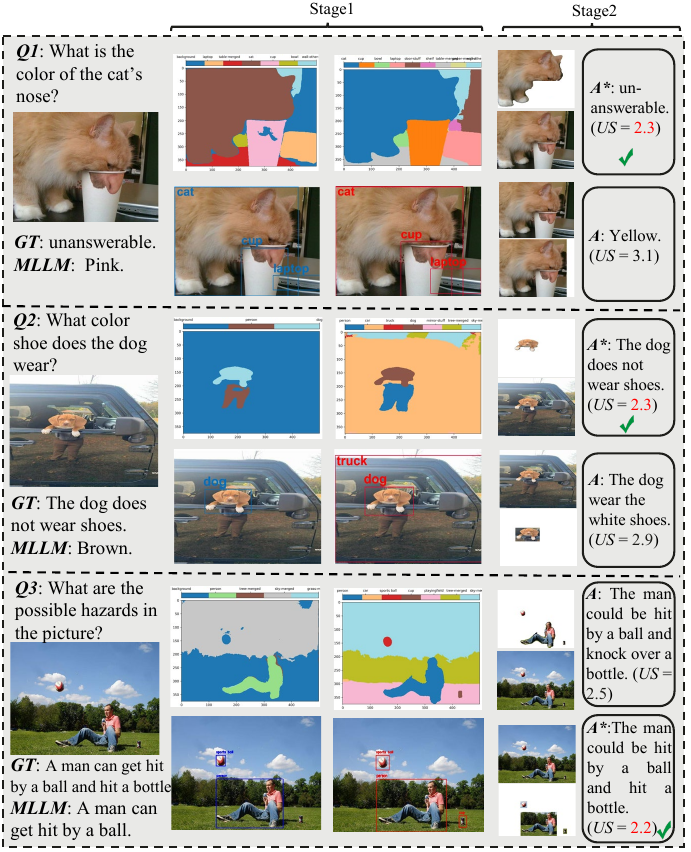}
    \vspace{-0.5cm}
    \caption{Visualization of some results. GT refers to the ground truth of the answer. Due to the limited space,  We do not show interaction with MLLM, focusing on the UQ  process in Stage 1 and Stage 2. Refer to Fig. \ref{fig:main_fig} for more details of the framework.}
    \vspace{-0.6cm}
    \label{fig:qualitativa_performance}
\end{figure}

We show some representative examples in Fig. \ref{fig:qualitativa_performance} to qualitatively analyze SRICE.  Due to the limited space, We omit the interaction with MLLM in the figure, mainly highlighting the results of the tool calibration and the answer choices for multiple pathways.  In Q1, the MLLM provides an incorrect answer due to interference from the cup pattern. The segmentation tool—especially after correction—accurately separates the cat from the cup, which helps the MLLM reach the correct answer with high confidence. Although the detection tool correctly identifies both the cup and the cat, its bounding box is too broad, failing to eliminate the interfering information. In Q2, the MLLM is confused by the image and produces an incorrect answer. In this case, both the segmentation and detection tools identify the dog; however, the segmentation tool also captures the person and refines the output through the calibration, which helps to better distinguish between the human and the dog and allows the MLLM to generate a more reliable answer. In Q3, MLLM fails to capture the bottle on the grass and gives an incomplete answer. The segmentation model and the detection model also face the same challenge. However, CP-based calibration enables the segmentation model to re-segment the bottle from the background, and the detection model successfully identifies this object. Consequently, MLLM focuses on the ball, the person, and the bottle, which leads to the correct answer. More visualizations are shown in Appendix C.

\vspace{-2mm}
\subsection{Ablation study} 
\vspace{-2mm}
We perform comprehensive ablation experiments to validate the effectiveness of the proposed approach. 

\noindent{\textbf{Different base model}}
We evaluated different base models in the SRICE framework, including mPLUG-Owl2 (7B) \cite{ye2024mplug} and Qwen-VL (7B) \cite{bai2023qwen}. When integrated with the SRICE framework, they are referred to as mPLUG-Owl2-SRICE and Qwen-VL-SRICE, respectively. As shown in Table \ref{tab:other_models}, the SRICE framework consistently enhances the performance of the base models across all datasets, with average improvements of 5.2$\%$ and 6.8$\%$ over five datasets, which demonstrates the generalizability of our proposed method.

\begin{table}[t]
\centering
\caption{\textcolor{black}{Comparison of different base models in SRICE.
}}
\label{tab:other_models}
\vspace{-0.3cm}
\centering
\setlength{\tabcolsep}{1pt} 
\resizebox{1\linewidth}{!}{
\begin{tabular}{lcccccc}
\toprule
            Method &  VQA2 &  VizWiz  &   GQA &  Flickr30K &  MMBench &  Average \\
\midrule
  mPLUG-Owl2& 78.3&	49.8&	56.1	&58.5&	64.5&	61.4
\\
 mPLUG-Owl2-SRICE &80.3&	56.9&	62.6	&63.8&	69.4&	66.6
 \\
Qwen-VL & 79.5&	42.6	&59.3	&56.0	&60.6&	59.6
 \\
Qwen-VL-SRICE & 81.9&	54.8&	65.7	&63.2&	66.2&	66.4
\\

\bottomrule
\end{tabular}
}
\vspace{-0.6cm}
\end{table}

\noindent{\textbf{Ablation of UQ approach in SRICE}}
We first compare the pipeline without using uncertainty to guide SRICE by disabling the calibration in Stage 1 and the uncertainty score calculation in Stage 2.  This yields two pathways: SRICE-seg-base (segmentation tool) and SRICE-det-base (object detection tool). Next, we enable tool calibration in Stage 1 on top of  SRICE-seg-base and SRICE-det-base to explore the benefit of the method, resulting in SRICE-seg-CP and SRICE-det-CP. The performances of these approaches are shown in Table \ref{tab:ablation_answer_selection}. By comparing SRICE-seg-CP, SRICE-det-CP, and SRICE, it shows that using quasi-conformal prediction for answer selection improves the performance of a single pathway, and SRICE outperforms any single-pathway method on all datasets. Compared to the strongest single pathway, SRICE achieves an average improvement of about 1$\%$ on five datasets. This suggests that different samples require different vision models and that the uncertainty in the MLLM output can serve as a reliable metric for choosing the more appropriate tool. 

\begin{figure}
    \centering
    \vspace{-0.3cm}
    \includegraphics[width=1\linewidth]{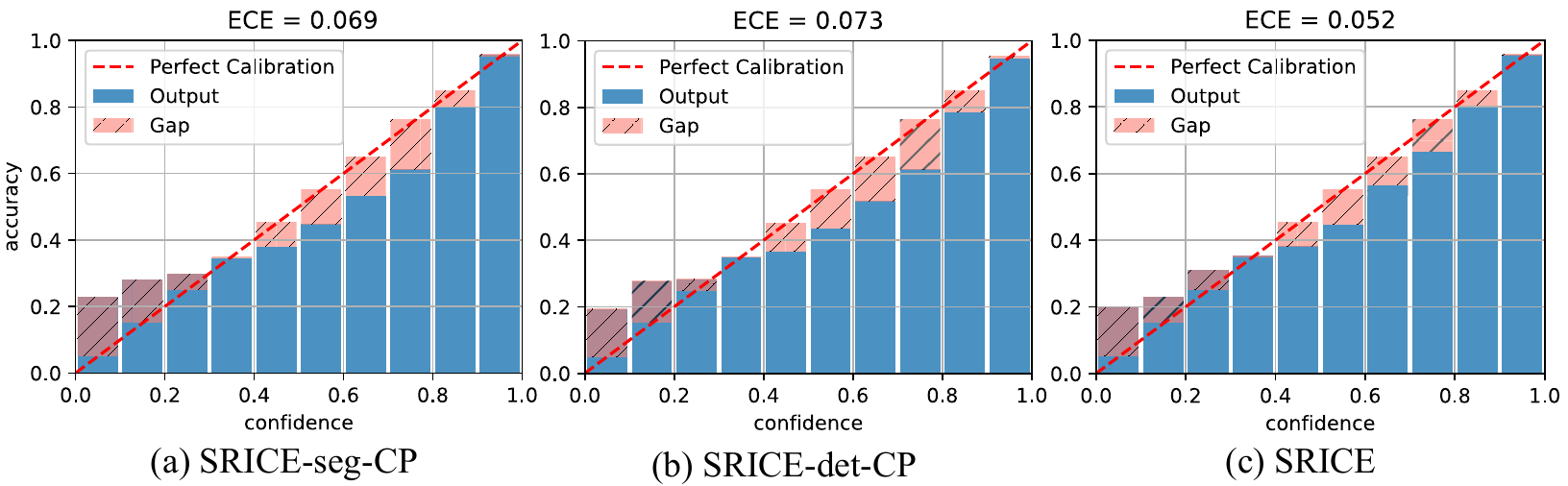}
    \vspace{-6mm}
    \caption{Reliability diagram of SRICE-seg-CP, SRICE-det-CP and SRICE on GQA dataset. }
    \label{fig:ECE}
    \vspace{-6mm}
\end{figure}

We show the reliability diagram of SRICE-seg-CP, SRICE-det-CP, and SRICE on the GQA dataset in Fig.~\ref{fig:ECE} to compare the reliability of the answer. A unique scores of 0 and 1 is calculated  for each answer in the GQA dataset. We normalize the uncertainty scores of all answers before subtracting them from one to obtain the confidence score. It shows that the expected calibration error (ECE) \cite{guo2017calibration} is reduced in SRICE, indicating the effectiveness of the proposed UQ for MLLM. Further comparison of SRICE-seg-base and SRICE-det-base with SRICE-seg-CP and SRICE-det-CP shows that CP-based tool calibration yields an improvement of 2.2$\%$ and 1.1$\%$ for two pathways, respectively. We attribute the higher performance of SRICE-seg-CP to its pixel-level ROI extraction, which excludes more irrelevant regions than box-level extraction and thus reduces the likelihood of introducing noise. 

\noindent{\textbf{Different UQ approaches in SRICE}} We compare different UQ approaches. For the external tools, we compare CP with a heuristic method. In the segmentation result, the confidence score for each object is obtained by averaging the softmax probabilities of all pixels belonging to it.  In the detection model, it uses the confidence score returned by the  model. We refer to these two pathways as SRICE-seg-CS and SRICE-det-CS, respectively, and the overall approach as SRICE-CS. To incorporate the confidence score, we include it in the tool’s return value  (e.g., \textit{'$\{$"id": xx, "name": 'xx', "confidence score": xx, "boundary": xx, ...... $\}$'}) and add the  text to $prompt_{roi}$: \textit{'xxx, Consider the semantic, location, size and confidence relationships 
between all objects and how they relate to the question. xxx'}. For MLLM's output, we compare with uncertainty estimated by entropy:
\begin{equation}
\small
US_{\text{entropy}}(A) = \frac{1}{T} \sum_{i=1}^{T} {\left( -\sum_{j=1}^{V} p_i^{(j)} \log p_i^{(j)} \right)}.
\end{equation}
We denote this approach as SRICE-US$\_$entropy, where the tool calibration still uses CP. Table \ref{tab:ablation_different_UQ} presents the results. From Table \ref{tab:ablation_different_UQ} we see that heuristic-based uncertainty estimation for the tools underperforms the CP-based approach in SRICE, with drops of 1.6$\%$ for the segmentation pathway and 0.8$\%$  for the detection pathway.  This effect is more pronounced in segmentation, likely because the heuristic struggles to handle overly conservative segmentation predictions. We also observe that using entropy for the MLLM output boosts performance by 0.4$\%$ compared to the strongest single pathway, but it still trails the quasi-conformal prediction approach by 0.6$\%$. We attribute it to  UQ by top-$p$ coverage focusing on major contributors to the probability mass and aligns more closely with practical uncertainty perceptions, whereas entropy factors in the entire long-tail distribution and can artificially inflate perceived uncertainty in some cases.

\noindent{\textbf{More tools in SRICE}} In principle, SRICE can incorporate any number of tools. To explore this, we add Mask2Former \cite{cheng2021mask2former} as a second segmentation tool (SRICE-seg2) and InternImage \cite{wang2023internimage} as a second detection tool (SRICE-det2). We refer to the extended approach  as SRICE+. Table \ref{tab:more_tools} shows the experimental results. We observe that SRICE-seg2-CP and SRICE-det2-CP perform similarly to the existing two pathways, and SRICE+ achieves only a modest 0.6$\%$ improvement. We speculate that the marginal improvement is because the fine-grained information provided by these tools is already of high quality after calibration. 
\begin{table}[t]
\centering
\caption{\textcolor{black}{Comprison of different UQ approaches. }}
\vspace{-0.3cm}
\resizebox{1\linewidth}{!}{
\setlength{\tabcolsep}{1pt} 
\begin{tabular}{lcccccc}
\toprule
            Method &  VQA2 &  VizWiz  &   GQA &  Flickr30K &  MMBench &  Average \\
\midrule
       
SRICE-seg-CP & 82.0 &   59.0 &66.6 &    66.1  &   75.5 &    69.8\\
SRICE-det-CP & 81.7 &   58.6 & 66.2 &    65.2 &  75.8 &    69.5 \\
SRICE-seg-CS &80.9	&57.1	&65.8	&63.0	&74.3	&68.2  \\
SRICE-det-CS &81.0	&57.6	&66.4	&63.4	&75.0	&68.7\\
SRICE-CS       &81.9	&58.3	&67.2	&64.0	&76.0	&69.5\\
SRICE-US$\_$entropy &82.0	&59.4	&66.9	&66.5	&76.3	&70.2 \\
SRICE & \textbf{82.5} &   \textbf{60.1} & \textbf{67.3} &    \textbf{67.2} &    \textbf{77.1} &             \textbf{70.8} \\
\bottomrule
\end{tabular}
}

\label{tab:ablation_different_UQ}
\vfill
\caption{\textcolor{black}{Performance of SRICE+ with 4 tools}}
\vspace{-0.3cm}
\centering
\setlength{\tabcolsep}{2pt} 
\resizebox{1\linewidth}{!}{
\begin{tabular}{lcccccc}
\toprule
            Method &  VQA2 &  VizWiz  &   GQA &  Flickr30K &  MMBench &  Average \\
\midrule
       
SRICE-seg-CP & 82.0 &   59.0 &66.6 &    66.1  &   75.5 &    69.8\\
SRICE-det-CP & 81.7 &   58.6 & 66.2 &    65.2 &  75.8 &    69.5 \\
SRICE-seg2-CP &81.5	&58.6	&66.0	&65.7	&75.0	&69.4  \\
SRICE-det2-CP &81.9	&59.1	&66.5	&65.6	&76.7	&70.0\\
SRICE & 82.5 &   60.1 & 67.3 &    67.2 &    77.1 &             70.8 \\
SRICE+ & \textbf{83.2}	&\textbf{60.7}	&\textbf{67.8}	&\textbf{67.8}	&\textbf{77.4}	&\textbf{71.4}\\
\bottomrule
\end{tabular}
}

\vspace{-0.6cm}
\label{tab:more_tools}
\end{table}

%% file: sec/5_conclusion.tex
\vspace{-2mm}
\section{Conclusion}\label{sec:con}
\vspace{-2mm}
In this work, we introduced SRICE, a training-free multimodal reasoning framework that leverages external vision tools alongside uncertainty quantification (UQ) based on conformal prediction (CP). Owing to its multi-stage structure, SRICE allows the MLLM to autonomously select regions of interest that guide the CoT process toward the final answer.  The CP-based UQ  not only provides a more reliable output from the vision tools but aids in selecting a more credible reasoning pathway.  Our approach addresses the challenges of boosting CoT in MLLM, which typically requires extensive labeling and training while mitigating the uncertainty introduced by tool-based methods.  Extensive experiments on five diverse datasets—VQA2, VizWiz, Flickr30K, GQA, and MMBench—demonstrate that SRICE consistently improves the base model’s performance, achieving an average improvement of 4.6$\%$ and, in some cases, even outperforming methods that require additional training. Ablation studies further confirm the benefits of our CP-based UQ, showing clear advantages over heuristic approaches.   In future work, we will explore extending SRICE to additional modalities.

%% file: sec/6_appendix.tex
\clearpage
\onecolumn
\setcounter{page}{1}
\setcounter{section}{0}
{\centering
        \Large
        \textbf{
        Appendix} \\
}
\renewcommand{\thesection}
{\Alph{section}} 
\section{Method}
We summarize the proposed SRICE framework in Algorithm \ref{alg:SRICE}.
\begin{algorithm}[ht]
\caption{The Proposed SRICE Framework}
\label{alg:SRICE}
\textbf{Input:} Image $I_0$, Question $Q_0$, external tools $f_{seg}$, $f_{det}$, pre-trained MLLM $f_{\text{MLLM}}$,   CP-based calibration methods ($CP_{seg}$, $CP_{det}$), the method for UQ of MLLM output   $US$, pixel-based image extraction $E_{seg}$ and bounding box-based image extraction $E_{det}$.
\begin{algorithmic}[1]
    \Statex \textbf{Stage 1: Tool Call and Calibration}
    \State Call segmentation tool: $R_{seg} \gets CP_{seg}(f_{seg}(I_0))$
    \State Call detection tool: $R_{det} \gets CP_{det}(f_{det}(I_0))$
    \State Select RoI via MLLM-CoT:
        \begin{align*}
            O_{seg} &\gets f_{MLLM}(R_{seg}, Q_0, prompt_{roi}), \\
            O_{det} &\gets f_{MLLM}(R_{det}, Q_0, prompt_{roi})
    \end{align*}
    \State Extract RoI images: 
    \[
        I_{seg}^{roi} = E_{seg}(I_0, O_{seg}),\quad I_{det}^{roi} = E_{det}(I_0, O_{det})
    \]

    \Statex \textbf{Stage 2: Final Answer Generation and Selection}
    \State Perform final reasoning with CoT:
    \[
        A_{seg} \gets f_{MLLM}(I_0, Q_0, I_{seg}^{roi}, prompt_{answer}), \quad
        A_{det} \gets f_{MLLM}(I_0, Q_0, I_{det}^{roi}, prompt_{answer})
    \]
    \State Compute uncertainty scores:
    \[
        US_{seg} = US(A_{seg}), \quad US_{det} = US(A_{det})
    \]
    \State Choose the final answer with minimal uncertainty:
    \[
        A^* = \arg\min_{A_k \in \{A_{seg}, A_{det}\}} US(A_k)
    \]
    \State \Return Final Answer: $A^*$
\end{algorithmic}
\end{algorithm}

\section{Experimental Settings}
\noindent{\textbf{Metric calculation method}}   All datasets  involved in the paper  use accuracy as an evaluation metric and calculate it as follows:
\begin{itemize}
    \item  VQA2. Each question in VQA2 is associated with 10 human-provided answers. For a predicted answer, the score is defined as: score = min(1, (number of matching human answers) / 3), which  means  if at least 3 annotators agree with the predicted answer, it receives full credit (a score of 1); fewer matches yield partial credit. The overall accuracy is the average of these per-question scores across the dataset.
    \item  VizWiz. Similar to VQA2, VizWiz collects multiple answers per question (typically reflecting the diversity in responses from blind users). It uses essentially the same formula as VQA2—comparing the predicted answer against the set of human responses using the min(1, count/3) rule. The per-question scores are averaged to produce the final accuracy score.
    \item  GQA.  For each question, the answer is deemed correct if it exactly matches the ground truth provided by the dataset. The primary measure is the percentage of questions answered correctly (i.e., the number of correct answers divided by the total number of questions). Beyond simple accuracy, GQA papers also explore secondary metrics (such as consistency, validity, and reasoning subtasks), but the headline metric remains overall accuracy.
    \item  Flickr30K. The evaluation of answers in Flickr30K follows \cite{shao2024visual}. Reference answers are provided, and predicted answers are evaluated against these references using a GPT-based scoring method. The final performance metric is computed by averaging the scores across all samples. 
    \item  MMbench. MMBench is designed to evaluate multimodal models over a range of tasks. For each question or task, answers are scored on a scale (often normalized between 0 and 1 or 0 and 100) based on criteria like correctness, completeness, and relevance. The scoring can be performed using human evaluations, automated assessments (e.g., leveraging language models as evaluators), or a combination of both. Each answer receives a score reflecting how well it meets the task requirements. These per-task (or per-question) scores are then averaged to yield a composite performance metric that summarizes the model’s overall capability across the diverse tasks included in MMBench.
\end{itemize}
\noindent{\textbf{Experimental platform}} All the experiments run on NVIDIA L40 GPU with CUDA 12.1. 
\section{More experimental results}
\begin{figure}
    \centering
    \includegraphics[width=0.8\linewidth]{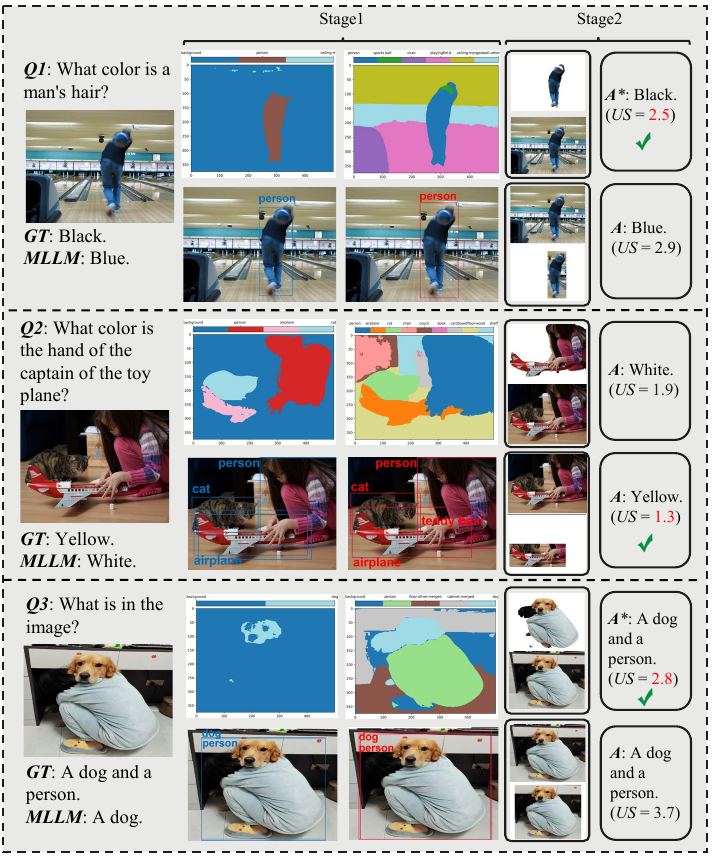}
    \caption{Visualization of more results.}
    \label{fig:qualitativa_performance_appendix}
\end{figure}

\noindent{\textbf{Visualization of more results}} We show more representative examples in Fig.\ref{fig:qualitativa_performance_appendix}. In Q1, the MLLM fails to distinguish between the bowling ball and the person’s head, leading to an incorrect answer. Initially, the segmentation tool identifies only a portion of the person’s region. However, after applying CP, it successfully segments both the bowling ball’s and the head’s regions. This improvement enables the model to capture the head’s details and mitigate the bowling ball’s interference, ultimately guiding the MLLM to the correct answer, whereas the detection model is unable to resolve the issue.  In Q2, note the airplane toy includes a \textbf{\textit{small captain toy on its front}}—a detail that the MLLM initially missed. Although neither the segmentation nor the detection model recognized this feature at first, the calibrated detection tool eventually identified an object in that place thereby preventing the model from focusing on the little girl, which assists the MLLM in arriving at the correct answer.   In Q3, the MLLM failed to distinguish between the human and the dog, resulting in an incorrect answer. The segmentation tool only segments the dog 
while a large area of the image is labeled as background. In 
contrast, the detection tool identifies both the human and the 
dog, though its detection frame lacked precision. After CP
calibration, the detection tool’s output changed little, but the 
segmentation model’s calibrated results successfully separated the human and the dog, enabling the MLLM to reach 
the correct answer with high confidence. In addition, combining Fig.  \ref{fig:qualitativa_performance} and Fig. \ref{fig:qualitativa_performance_appendix}, we observe that the segmentation model performs better with overlapping targets, whereas the bounding boxes provided by the detection model struggle to accurately separate these targets. 

\noindent{\textbf{Different $\alpha$ and $p$ in SRICE framework}} We evaluate the effects of parameters $\alpha$ and $p$ in the SRICE framework by conducting experiments with varying settings. Specifically, we examine $\alpha$ values at 0.05, 0.1, 0.15, and 0.2, and $p$ values at 0.8, 0.85, 0.9, and 0.95. The performance results are summarized in Table \ref{tab:different_alpha} and Table \ref{tab:different_p}. From Table \ref{tab:different_alpha}, we observe that extreme values of $\alpha$ negatively impact the performance. Specifically, very small $\alpha$ values impose stringent error-rate constraints, resulting in wider, overly conservative prediction intervals that lack granularity and informativeness. Conversely, large $\alpha$ values permit higher error rates, yielding narrower, more precise intervals but significantly increasing the likelihood that the true value falls outside the interval. Table \ref{tab:different_p} indicates that SRICE performance remains relatively stable when $p$ is set to 0.95. However, reducing $p$ to 0.85 or 0.8 leads to noticeable performance degradation. This decline likely occurs because a lower threshold prematurely discards predictions with closely matched probabilities, thereby truncating the probability distribution and misrepresenting the actual uncertainty. Based on our analysis, we recommend selecting $\alpha = 0.1$ and $p = 0.9$ for optimal performance of SRICE.

\begin{table*}[h]
\caption{\textcolor{black}{The performance of SRICE with different $\alpha$. }}
\label{tab:different_alpha}
\centering
\begin{tabular}
{lcccccc}
\toprule
        $\alpha$ &  VQA2 &  VizWiz  &   GQA &  Flickr30K &  MMBench &  Average \\
\midrule
0.05 & 82.1	&59.5	&67.0	&66.6&	77.0&	70.4
 \\
0.1 (\textit{used in paper})  & \textbf{82.5} &\textbf{ 60.1} &\textbf{ 67.3} &\textbf{ 67.2} &    \textbf{77.1} & \textbf{70.8} \\
 0.15 &82.3	&59.8	&66.8&	66.9&	76.5&	70.5\\
  0.2 & 82.0	&59.2	&66.6&	66.5&	76.2	&70.1
 \\
\bottomrule
\end{tabular}
\end{table*}
 
\begin{table*}[h]
\caption{\textcolor{black}{The performance of SRICE with different $p$. }}
\label{tab:different_p}
\centering
\begin{tabular}
{lcccccc}
\toprule
        $p$ &  VQA2 &  VizWiz  &   GQA &  Flickr30K &  MMBench &  Average \\
\midrule
0.8 & 81.8& 	59.0& 	66.5& 	65.9	& 76.2	& 69.9

 \\
0.85  & 82.0& 	59.3	& 66.8& 	66.6& 	76.7	& 70.3
\\
 0.9 (\textit{used in paper}) & \textbf{82.5} &\textbf{ 60.1} &\textbf{ 67.3} &\textbf{ 67.2} &    \textbf{77.1} & \textbf{70.8}\\
  0.95 & 82.5& 	59.8& 	67.0	& 67.0	& 77.0& 	70.7

 \\
\bottomrule
\end{tabular}
\end{table*}

